%% file: ms.tex
\documentclass[11pt,a4paper]{article}
\usepackage[hyperref]{emnlp-ijcnlp-2019}
\usepackage{times}
\usepackage{amsmath}
\usepackage{latexsym}
\usepackage{graphicx}
\usepackage{tabularx}
\usepackage{url}
\usepackage{cleveref}
\usepackage[inline]{enumitem}
\usepackage{array}
\usepackage{setspace}
\usepackage{multirow}
\usepackage{hyperref}
\usepackage{framed}
\usepackage{color}
\usepackage{xcolor}
\usepackage{xspace}
\usepackage{amssymb}
\usepackage{booktabs}
\usepackage{colortbl}
\usepackage{txfonts}

\DeclareMathOperator{\softmax}{softmax}

\newcommand{\eg}{\textit{e.g.}}
\newcommand{\ie}{\textit{i.e.}}
\newcommand{\mask}{\textsc{[Mask]}}
\newcommand{\rel}[1]{\verb~#1~}
\newcommand{\ent}[1]{\textsc{#1}}

\aclfinalcopy 

\newcommand{\fair}{$^1$}
\newcommand{\ucl}{$^2$}
\newcommand{\uclfair}{$^{1,2}$}

\title{Language Models as Knowledge Bases?}

\author{
Fabio Petroni\fair{} \ Tim Rockt\"aschel\uclfair{} \  \ Patrick Lewis\uclfair{} \ Anton Bakhtin\fair{}  \\
{\bf   Yuxiang Wu\uclfair{} \ Alexander H. Miller\fair{} \ Sebastian Riedel\uclfair{} }\\
\fair{}Facebook AI Research \\
\ucl{}University College London\\
{\tt \{fabiopetroni, rockt, plewis, yolo, yuxiangwu, ahm, sriedel\}@fb.com}
}

\date{}

\begin{document}

\maketitle
\begin{abstract}
  \input{0_abstract.tex}
\end{abstract}

\input{1_introduction.tex}
\input{2_background.tex}
\input{3_related_work.tex}
\input{4_methods-new.tex}
\input{5_results.tex}
\input{6_conclusion.tex}

\section*{Acknowledgments}
We would like to thank the reviewers for their thoughtful comments and efforts towards improving our manuscript. In addition, we would like to acknowledge three frameworks that were used in our experiments: AllenNLP\footnote{\url{https://github.com/allenai/allennlp}}, Fairseq\footnote{\url{https://github.com/pytorch/fairseq}} and the Hugging Face PyTorch-Transformers\footnote{\url{https://github.com/huggingface/pytorch-transformers}} library.

\bibliography{acl2019,lama,rockt,riedel}
\bibliographystyle{acl_natbib}

\end{document}

%% file: 0_abstract.tex
Recent progress in pretraining language models on large textual corpora led to a surge of improvements for downstream NLP tasks.
Whilst learning linguistic knowledge, these models may also be storing relational knowledge present in the training data, and may be able to answer queries structured as ``fill-in-the-blank" cloze statements.
Language models have many advantages over structured knowledge bases: they require no schema engineering, allow practitioners to query about an open class of relations, are easy to extend to more data, and require no human supervision to train.
We present an in-depth analysis of the relational knowledge already present (without fine-tuning) in a wide range of state-of-the-art pretrained language models.
We find that \begin{enumerate*}[label=(\roman*)]
\item without fine-tuning, BERT contains relational knowledge competitive with traditional NLP methods that have some access to oracle knowledge,
\item BERT also does remarkably well on open-domain question answering against a supervised baseline, and
\item certain types of factual knowledge are learned much more readily than others by standard language model pretraining approaches.
\end{enumerate*}
The surprisingly strong ability of these models to recall factual knowledge without any fine-tuning demonstrates their potential as unsupervised open-domain QA systems. 
The code to reproduce our analysis is available at
\url{https://github.com/facebookresearch/LAMA}.

%% file: 1_introduction.tex
\section{Introduction}

\begin{figure}
    \centering
     \includegraphics[width=\columnwidth]{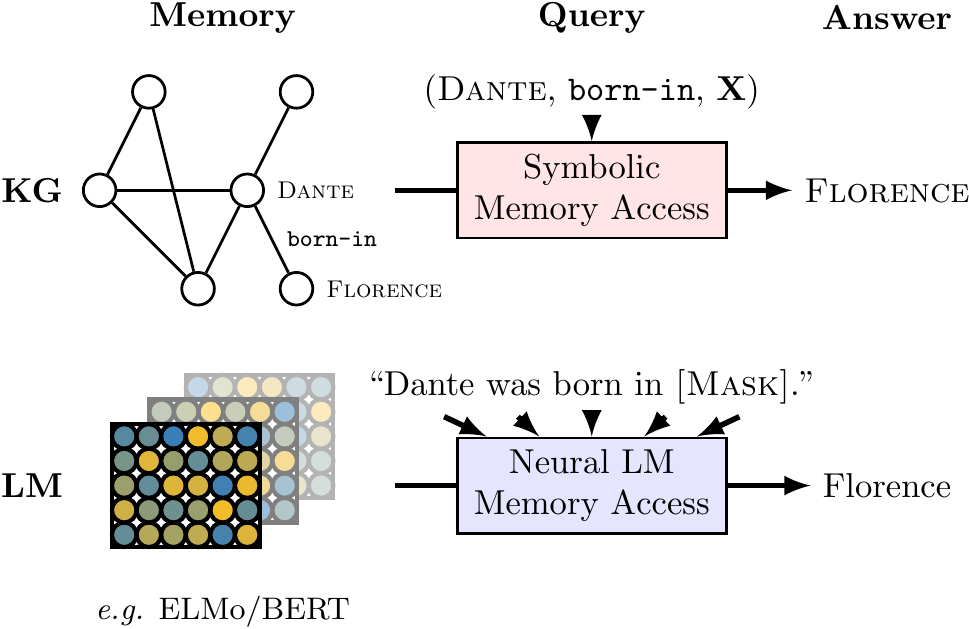}  
    \caption{Querying knowledge bases (KB) and language models (LM) for factual knowledge.
    }
    \label{fig:overview}
\end{figure}

Recently, pretrained high-capacity language models such as ELMo \citep{DBLP:conf/naacl/PetersNIGCLZ18} and BERT \citep{DBLP:journals/corr/abs-1810-04805} have become increasingly important in NLP. 
They are optimised to either predict the next word in a sequence or some masked word anywhere in a given sequence (\eg{} ``Dante was born in \mask{} in the year 1265.'').
The parameters of these models appear to store vast amounts of linguistic knowledge \citep{DBLP:conf/emnlp/PetersNZY18,DBLP:journals/corr/abs-1901-05287,tenney2018what} useful for downstream tasks.
This knowledge is usually accessed either by conditioning on latent context representations produced by the original model or by using the original model weights to initialize a task-specific model which is then further fine-tuned.
This type of knowledge transfer is crucial for current state-of-the-art results on a wide range of tasks.

In contrast, knowledge bases are effective solutions for accessing annotated gold-standard relational data by enabling queries such as (\ent{Dante}, \rel{born-in}, $\mathbf{X}$). However, in practice we often need to \emph{extract} relational data from text or other modalities to populate these knowledge bases. This requires complex NLP pipelines involving entity extraction, coreference resolution, entity linking and relation extraction~\cite{surdeanu_overview_2014}---components that often need supervised data and fixed schemas.
Moreover, errors can easily propagate and accumulate throughout the pipeline.
Instead, we could attempt to query neural language models for relational data by asking them to fill in masked tokens in sequences like ``Dante was born in \mask{}'', as illustrated in \Cref{fig:overview}. In this setting, language models come with various attractive properties: they require no schema engineering, do not need human annotations, and they support an open set of queries.

Given the above qualities of language models as potential representations of relational knowledge, we are interested in the relational knowledge already present in pretrained \emph{off-the-shelf} language models such as ELMo and BERT. 
How much relational knowledge do they store? How does this differ for different types of knowledge such as facts about entities, common sense, and general question answering? How does their performance without fine-tuning compare to symbolic knowledge bases automatically extracted from text? Beyond gathering a better general understanding of these models, we believe that answers to these questions can help us design better unsupervised knowledge representations that could transfer factual and commonsense knowledge reliably to downstream tasks such as commonsense (visual) question answering \citep{DBLP:journals/corr/abs-1811-10830,DBLP:conf/naacl/TalmorHLB19} or reinforcement learning \citep{branavan2011learning,DBLP:journals/corr/abs-1810-08272,bahdanau2019learning,luketina2019survey}.

For the purpose of answering the above questions we introduce the \emph{LAMA} (LAnguage Model  Analysis) probe, consisting of a set of knowledge sources, each comprised of a set of facts. We define that a pretrained language model \emph{knows} a fact (\ent{subject}, \rel{relation}, \ent{object}) such as (\ent{Dante}, \rel{born-in}, \ent{Florence}) if it can successfully predict masked  objects in cloze sentences such as ``Dante was born in \underline{\hspace{2em}}''  expressing that fact. We test for a variety of types of knowledge: relations between entities stored in Wikidata, common sense relations between concepts from ConceptNet, and knowledge necessary to answer natural language questions in SQuAD. In the latter case we manually map a subset of SQuAD questions to cloze sentences.

Our investigation reveals that
\begin{enumerate*}[label=(\roman*)] 
    \item the largest BERT model from \citet{devlin_bert:_2018} (BERT-large) captures (accurate) relational knowledge comparable to that of a knowledge base extracted with an off-the-shelf relation extractor and an oracle-based entity linker from a corpus known to express the relevant knowledge,  
    \item factual knowledge can be recovered surprisingly well from pretrained language models, however, for some relations (particularly $N$-to-$M$ relations) performance is very poor,
    \item BERT-large consistently outperforms other language models in recovering factual and commonsense knowledge while at the same time being more robust to the phrasing of a query, and
    \item BERT-large achieves remarkable results for open-domain QA, reaching 57.1\% precision@10 compared to 63.5\% of a knowledge base constructed using a task-specific supervised relation extraction system  
\end{enumerate*}.

%% file: 2_background.tex
\section{Background}
\label{sec:Background}

\begin{table*}[h!]
    \centering
\resizebox{\textwidth}{!}{    
    \begin{tabular}{lcccc}
        \toprule
        Model &  Base Model & \#Parameters & Training Corpus & Corpus Size\\
        \midrule
        fairseq-fconv \citep{DBLP:conf/icml/DauphinFAG17}
 & ConvNet & 324M & WikiText-103 & 103M Words\\
        Transformer-XL (large) \citep{DBLP:journals/corr/abs-1901-02860} & Transformer & 257M & WikiText-103 & 103M Words\\
        \midrule
        ELMo (original) \citep{DBLP:conf/naacl/PetersNIGCLZ18} & BiLSTM & 93.6M  & Google Billion Word & 800M Words\\ 
        ELMo 5.5B \citep{DBLP:conf/naacl/PetersNIGCLZ18} & BiLSTM & 93.6M & Wikipedia (en) \& WMT 2008-2012 & 5.5B Words\\ 
        BERT (base) \citep{DBLP:journals/corr/abs-1810-04805} & Transformer & 110M & Wikipedia (en)  \& BookCorpus& 3.3B Words \\
        BERT (large) \citep{DBLP:journals/corr/abs-1810-04805} & Transformer & 340M & Wikipedia (en)  \& BookCorpus & 3.3B Words \\     
        \bottomrule
    \end{tabular}
}
    \caption{Language models considered in this study.}
    \label{tab:models}
\end{table*}

In this section we provide background on language models. Statistics for the models that we include in our investigation are summarized in \Cref{tab:models}.

\subsection{Unidirectional Language Models}
Given an input sequence of tokens $\mathbf{w} = [w_1, w_2, \ldots, w_N]$, unidirectional language models commonly assign a probability $p(\mathbf{w})$ to the sequence by factorizing it as follows
\begin{equation}
    p(\mathbf{w}) = \prod_t p(w_t\,|\,w_{t-1}, \ldots, w_1).
\end{equation}
A common way to estimate this probability is using neural language models~\citep{DBLP:conf/slt/MikolovZ12,DBLP:journals/corr/MelisDB17,DBLP:journals/jmlr/BengioDVJ03} with
\begin{equation}
    p(w_t\,|\,w_{t-1}, \ldots, w_1) = \softmax(\mathbf{W}\mathbf{h}_t+\mathbf{b})
    \label{eq:toplayer}
\end{equation}
where $\mathbf{h}_t \in \mathbb{R}^k$ is the output vector of a neural network at position $t$ and $\mathbf{W} \in \mathbb{R}^{|\mathcal{V}|\ \times\ k}$ is a learned parameter matrix that maps $\mathbf{h}_t$ to unnormalized scores for every word in the vocabulary $\mathcal{V}$.
Various neural language models then mainly differ in how they compute $\mathbf{h}_t$ given the word history, \eg{}, by using a multi-layer perceptron \citep{DBLP:journals/jmlr/BengioDVJ03,DBLP:conf/slt/MikolovZ12}, convolutional layers \citep{DBLP:conf/icml/DauphinFAG17}, recurrent neural networks \citep{zaremba2014recurrent,DBLP:journals/corr/MerityXBS16,DBLP:journals/corr/MelisDB17} or self-attention mechanisms \citep{radford2018improving,DBLP:journals/corr/abs-1901-02860,radford2019language}.

\noindent
\textbf{fairseq-fconv}:
Instead of commonly used recurrent neural networks, \citet{DBLP:conf/icml/DauphinFAG17} use multiple layers of gated convolutions. We use the pretrained model in the fairseq\footnote{\url{https://github.com/pytorch/fairseq}} library in our study.
It has been trained on the WikiText-103 corpus introduced by \citet{DBLP:journals/corr/MerityXBS16}.

\noindent
\textbf{Transformer-XL}:
\citet{DBLP:journals/corr/abs-1901-02860} introduce a large-scale language model based on the Transformer \citep{DBLP:conf/nips/VaswaniSPUJGKP17}.
Transformer-XL can take into account a longer history by caching previous outputs and by using relative instead of absolute positional encoding. It achieves a test perplexity of $18.3$ on the WikiText-103 corpus.

\subsection{Bidirectional ``Language Models''\footnote{Contextual representation models \citep{tenney2018what} might be a better name, but we keep calling them language models for simplicity.}}
So far, we have looked at language models that predict the next word given a history of words. 
However, in many downstream applications we mostly care about having access to contextual representations of words, \ie{}, word representations that are a function of the entire context of a unit of text such as a sentence or paragraph, and not only conditioned on previous words.
Formally, given an input sequence $\mathbf{w} = [w_1, w_2, \ldots, w_N]$ and a position $1 \leq i \leq N$, we want to estimate $p(w_i) = p(w_i\,|\,w_1, \ldots, w_{i-1}, w_{i+1}, \ldots, w_N)$ using the left and right context of that word.

\noindent
\textbf{ELMo}:
To estimate this probability, \citet{DBLP:conf/naacl/PetersNIGCLZ18} propose running a forward and backward LSTM \citep{hochreiter1997long}, resulting in $\overrightarrow{\mathbf{h}}_i$ and $\overleftarrow{\mathbf{h}}_i$ which consequently are used to calculate a forward and backward language model log-likelihood.
Their model, ELMo, uses multiple layers of LSTMs and it has been pretrained on the Google Billion Word dataset.
Another version of the model, ELMo 5.5B, has been trained on the English Wikipedia and monolingual news crawl data from WMT 2008-2012.

\noindent
\textbf{BERT}:
Instead of a standard language model objective, \citet{DBLP:journals/corr/abs-1810-04805} propose to sample positions in the input sequence randomly and to learn to fill the word at the masked position. 
To this end, they employ a Transformer architecture and train it on the BookCorpus \citep{DBLP:conf/iccv/ZhuKZSUTF15} as well as a crawl of  English Wikipedia.
In addition to this pseudo language model objective, they use an auxiliary binary classification objective to predict whether a particular sentence follows the given sequence of words.

%% file: 3_related_work.tex
\section{Related Work}
Many studies have investigated pretrained word representations, sentence representations, and language models. Existing work focuses on understanding linguistic and semantic properties of word representations or how well pretrained sentence representations and language models transfer linguistic knowledge to downstream tasks.
In contrast, our investigation seeks to answer to what extent pretrained language models store factual and commonsense knowledge by comparing them with symbolic knowledge bases populated by traditional relation extraction approaches.

\citet{DBLP:conf/acl/BaroniDK14} present a systematic comparative analysis between neural word representation methods and more traditional count-based distributional semantic methods on lexical semantics tasks like semantic relatedness and concept categorization. They find that neural word representations outperform count-based distributional methods on the majority of the considered tasks.
\citet{DBLP:journals/coling/HillRK15} investigate to what degree word representations capture semantic meaning as measured by similarity between word pairs.

\citet{DBLP:conf/emnlp/MarvinL18} assess the grammaticality of pretrained language models.
Their dataset consists of sentence pairs with a grammatical and an ungrammatical sentence. 
While a good language model should assign higher probability to the grammatical sentence, they find that LSTMs do not learn syntax well.

Another line of work investigates the ability of pretrained sentence and language models to transfer knowledge to downstream natural language understanding tasks \citep{DBLP:conf/emnlp/WangSMHLB18}.
While such an analysis sheds light on the transfer-learning abilities of pretrained models  for understanding short pieces of text, it provides little insight into whether these models can compete with traditional approaches to representing knowledge like symbolic knowledge bases.

More recently, \citet{mccoy2019right} found that for natural language inference, a model based on BERT learns to rely heavily on fallible syntactic heuristics instead of a deeper understanding of the natural language input.
\citet{DBLP:conf/emnlp/PetersNZY18} found that lower layers in ELMo specialize on local syntactic relationships, while higher layers can learn to model long-range relationships. 
Similarly, \citet{DBLP:journals/corr/abs-1901-05287} found that BERT captures English syntactic phenomena remarkably well.
\citet{tenney2018what} investigate to what extent language models encode sentence structure for different syntactic and semantic phenomena and found that they excel for the former but only provide small improvements for tasks that fall into the latter category.
While this provides insights into the linguistic knowledge of language models, it does not provide insights into their factual and commonsense knowledge.

\citet{radford2018improving} introduce a pretrained language model based on the Transformer which they termed generative pretraining (GPTv1).
The first version of GPT \citep{radford2018improving} has been trained on the Book Corpus \citep{DBLP:conf/iccv/ZhuKZSUTF15} containing 7000 books.
The closest to our investigation is the work by \citet{radford2019language} which introduces GPTv2 and investigates how well their language model does zero-shot transfer to a range of downstream tasks.
They find that GPTv2 achieves an $F_1$ of 55 for answering questions in CoQA \citep{DBLP:journals/corr/abs-1808-07042} and $4.1\%$ accuracy on the Natural Questions dataset \citep{kwiatkowski2019natural}, in both cases without making use of annotated question-answer pairs or an information retrieval step.
While these results are encouraging and hint at the ability of very large pretrained language models to memorize factual knowledge, the large GPTv2 model has not been made public and the publicly available small version achieves less than $1\%$ on Natural Questions (5.3 times worse than the large model).
Thus, we decided to not include GPTv2 in our study.
Similarly, we do not include GPTv1 in this study as it uses a limited lower-cased vocabulary, making it incompatible to the way we assess the other language models.

%% file: 4_methods-new.tex
\section{The LAMA Probe}
\label{sec:method}

We introduce the \emph{LAMA} (LAnguage Model  Analysis) probe to test the factual and commonsense knowledge in language models.
It provides a set of knowledge sources which are composed of a corpus of facts.
Facts are either subject-relation-object triples or question-answer pairs. 
Each fact is converted into a cloze statement which is used to query the language model for a missing token. 
We evaluate each model based on how highly it ranks the ground truth token against every other word in a fixed candidate vocabulary. This is similar to ranking-based metrics from the knowledge base completion literature \citep{bordes2013translating,nickel2016review}.
Our assumption is that models which rank ground truth tokens high for these cloze statements have more factual knowledge.
We discuss each step in detail next and provide considerations on the probe below.

\subsection{Knowledge Sources}
To assess the different language models in \Cref{sec:Background}, we cover a variety of sources of factual and commonsense knowledge.
For each source, we describe the origin of fact triples (or question-answer pairs), how we transform them into cloze templates, and to what extent aligned texts exist in Wikipedia that are known to express a particular fact. 
We use the latter information in supervised baselines that extract knowledge representations directly from the aligned text. 

\subsubsection{Google-RE}
The Google-RE corpus\footnote{\url{https://code.google.com/archive/p/relation-extraction-corpus/}} contains ${\sim}60$K facts  manually extracted from Wikipedia. It covers five relations but we consider only three of them, namely ``place of birth'', ``date of birth'' and ``place of death''. We exclude the other two because they contain mainly multi-tokens objects that are not supported in our evaluation.
We manually define a template for each considered relation, \eg{}, ``[S] was born in [O]'' for  ``place of birth''.
Each fact in the Google-RE dataset is, by design, manually aligned to a short piece of Wikipedia text supporting it. 

\subsubsection{T-REx}
The T-REx knowledge source is a subset of Wikidata triples. It is derived from the T-REx dataset~\cite{elsahar2018t} and is much larger than Google-RE with a broader set of relations. 
We consider 41 Wikidata relations and subsample at most 1000 facts per relation.
As with the Google-RE corpus, we manually define a template for each relation (see Table \ref{tab:trexexamplesbert} for some examples). 
In contrast to the Google-RE knowledge source, T-REx facts were automatically aligned to Wikipedia and hence this alignment can be noisy. 
However, \citet{elsahar2018t} report an accuracy
of 97.8\% for the alignment technique over a test set. 

\subsubsection{ConceptNet}

ConceptNet~\cite{speer2012representing} is a multi-lingual knowledge base, initially built on top of Open Mind Common Sense (OMCS) sentences. 
OMCS represents commonsense relationships between words and/or phrases.
We consider facts from the English part of ConceptNet that have single-token objects covering 16 relations.
For these ConceptNet triples, we find the OMCS sentence that contains both the subject and the object.
We then mask the object within the sentence and use the sentence as template for querying language models. 
If there are several sentences for a triple, we pick one at random. 
Note that for this knowledge source there is no explicit alignment of facts to Wikipedia sentences. 

\subsubsection{SQuAD}
SQuAD \cite{rajpurkar_squad:_2016} is a popular question answering dataset. We select a subset of 305 context-insensitive questions from the SQuAD development set with single token answers.
We manually create cloze-style questions from these questions, \eg{}, rewriting ``Who developed the theory of relativity?'' as ``The theory of relativity was developed by \underline{\hspace{2em}}''.
For each question and answer pair, we know that the corresponding fact is expressed in Wikipedia since this is how SQuAD was created.

\subsection{Models}

We consider the following pretrained case-sensitive language models in our study (see Table \ref{tab:models}): fairseq-fconv (\textit{Fs}), Transformer-XL large (\textit{Txl}),  ELMo original (\textit{Eb}), ELMo 5.5B (\textit{E5B}), BERT-base (\textit{Bb}) and BERT-large (\textit{Bl}).
We use the natural way of generating tokens for each model by following the definition of the training objective function.

Assume we want to compute the generation for the token at position $t$.
For unidirectional language models, we use the network output ($\mathbf{h}_{t-1}$) just before the token to produce the output layer softmax.
For ELMo we consider the output just before ($\overrightarrow{\mathbf{h}}_{t-1}$) for the forward direction and just after ($\overleftarrow{\mathbf{h}}_{t+1}$) for the backward direction. Following the loss definition in \citep{DBLP:conf/naacl/PetersNIGCLZ18}, we average forward and backward probabilities from the corresponding softmax layers.
For BERT, we mask the token at position $t$, and we feed the output vector corresponding to the masked token ($\mathbf{h}_{t}$) into the softmax layer.
To allow a fair comparison, we let models generate over a unified vocabulary, which is the intersection of the vocabularies for all considered models (${\sim}21$K case-sensitive tokens).

\subsection{Baselines}
To compare language models to canonical ways of using off-the-shelf systems for extracting symbolic knowledge and answering questions, we consider the following baselines.

\noindent
\textbf{Freq}: For a subject and relation pair, this baseline ranks words based on how frequently they appear as objects for the given relation in the test data. It indicates the upper bound performance of a model that always predicts the same objects for a particular relation.

\noindent
\textbf{RE}: For the relation-based knowledge sources, we consider the pretrained Relation Extraction (RE) model of \citet{sorokin2017context}. This model was trained on a subcorpus of Wikipedia annotated with Wikidata relations. It extracts relation triples from a given sentence using an LSTM-based encoder and an attention mechanism. Based on the alignment information from the knowledge sources, we provide the relation extractor with the sentences known to express the test facts. 
Using these datasets, RE constructs a knowledge graph of triples. 
At test time, we query this graph by finding the subject entity and then rank all objects in the correct relation based on the confidence scores returned by RE.   
We consider two versions of this procedure that differ in how the entity linking is implemented:
\textbf{RE$_n$} makes use of a na\"ive entity linking solution based on exact string matching, while
\textbf{RE$_o$} uses an oracle for entity linking in addition to string matching. In other words, assume we query for the object $o$ of a test subject-relation fact $(s,r,o)$ expressed in a sentence $x$. If RE has extracted any triple $(s',r,o')$ from that sentence $x$, $s'$ will be linked to $s$ and $o'$ to $o$. In practice, this means RE can return the correct solution $o$ if \emph{any} relation instance of the right type was extracted from $x$, regardless of whether it has a wrong subject or object.

\noindent
\textbf{DrQA}: \citet{DBLP:journals/corr/ChenFWB17} introduce DrQA, a popular system for open-domain question answering. DrQA predicts answers to natural language questions using a two step pipeline. 
First, a TF/IDF information retrieval step is used to find relevant articles from a large store of documents (e.g. Wikipedia).
On the retrieved top $k$ articles, a neural reading comprehension model then extracts answers. 
To avoid giving the language models a competitive advantage, we constrain the predictions of DrQA to single-token answers.

\subsection{Metrics}
We consider rank-based metrics and compute results per relation along with mean values across all relations. To account for multiple valid objects for a subject-relation pair (\ie{}, for N-M relations), we follow \citet{bordes2013translating} and remove from the candidates when ranking at test time all other valid objects in the training data other than the one we test.
We use the mean precision at k (\textit{P@k}). For a given fact, this value is 1 if the object is ranked among the top k results, and 0 otherwise.

\subsection{Considerations}
There are several important design decisions we made when creating the LAMA probe. Below we give more detailed justifications for these decisions.

\paragraph{Manually Defined Templates}
For each relation we manually define a template that queries for the object slot in that relation. One can expect that the choice of templates has an impact on the results, and this is indeed the case: for some relations we find both worse and better ways to query for the same information (with respect to a given model) by using an alternate template. We argue that this means we are measuring a \emph{lower} bound for what language models know. We make this argument by analogy with traditional knowledge bases: they only have a \emph{single} way of querying knowledge for a specific relation, namely by using the relation id of that relation, and this way is used to measure their accuracy. For example, if the relation ID is \rel{works-For} and the user asks for \rel{is-working-for}, the accuracy of the KG would be 0.

\paragraph{Single Token}

We only consider single token objects as our prediction targets. 
The reason we include this limitation is that multi-token decoding adds a number of additional tuneable parameters (beam size, candidate scoring weights, length normalization, n-gram repetition penalties, etc.) that obscure the knowledge we are trying to measure. Moreover, well-calibrated multi-token generation is still an active research area, particularly for bidirectional models (see \eg{} \citet{welleck2019textgen}).

\paragraph{Object Slots}
We choose to \emph{only} query object slots in triples, as opposed to subject or relation slots. By including reverse relations (e.g. \rel{contains} and \rel{contained-by}) we can also query subject slots. We do not query relation slots for two reasons. First, surface form realisations of relations will span several tokens, and as we discussed above, this poses a technical challenge that is not in the scope of this work. Second, even if we could easily predict multi-token phrases, relations can generally be expressed with many different wordings, making it unclear what the gold standard pattern for a relation should be, and how to measure accuracy in this context.

\paragraph{Intersection of Vocabularies}
The models that we considered are trained with different vocabularies. 
For instance, ELMo uses a list of $\sim$800K tokens while BERT considers only $\sim$30K tokens. 
The size of the vocabulary can influence the performance of a model for the LAMA probe. 
Specifically, the larger the vocabulary the harder it would be to rank the gold token at the top. For this reason we considered a common vocabulary of $\sim$21K case-sensitive tokens that are obtained from the intersection of the vocabularies for all considered models. To allow a fair comparison, we let every model rank only tokens in this joint vocabulary.

%% file: 5_results.tex
\input{tables/TABLE2.tex}
\section{Results}
We summarize the main results in
\Cref{tab:templatep1}, which shows the mean precision at one (P@1) for the different models across the set of corpora considered.
In the remainder of this section, we discuss the results for each corpus in detail.

\paragraph{Google-RE}

We query the LMs using a standard cloze template for each relation.
The base and large versions of BERT both outperform all other models by a substantial margin. 
Furthermore, they obtain a $2.2$ and $2.9$ respective average accuracy improvement over the oracle-based RE baseline. 
This is particularly surprising given that with the gold-aligned Google-RE source we know for certain that the oracle RE baseline has seen at least one sentence expressing each test fact. 
Moreover, the RE baseline was given substantial help through an entity linking oracle.

It is worth pointing out that while BERT-large does better, this does not mean it does so for the right reasons. Although the aligned Google-RE sentences are likely in its training set (as they are part of Wikipedia and BERT has been trained on Wikipedia), it might not ``understand'' them to produce these results. 
Instead, it could have learned associations of objects with subjects from co-occurrence patterns.

\paragraph{T-REx}
The knowledge source derived from Google-RE contains relatively few facts and only three relations.
Hence, we perform experiments on the larger set of facts and relations in T-REx. 
We find that results are generally consistent with Google-RE. 
Again, the performance of BERT in retrieving factual knowledge are close to the performance obtained by automatically building a knowledge base with an off-the-shelf relation extraction system and oracle-based entity linking. 
Broken down by relation type, the performance of BERT is very high for 1-to-1 relations (\eg{}, \textit{capital of}) and low for N-to-M relations.

\begin{figure}[!t]
    \centering
    \includegraphics[width=\linewidth]{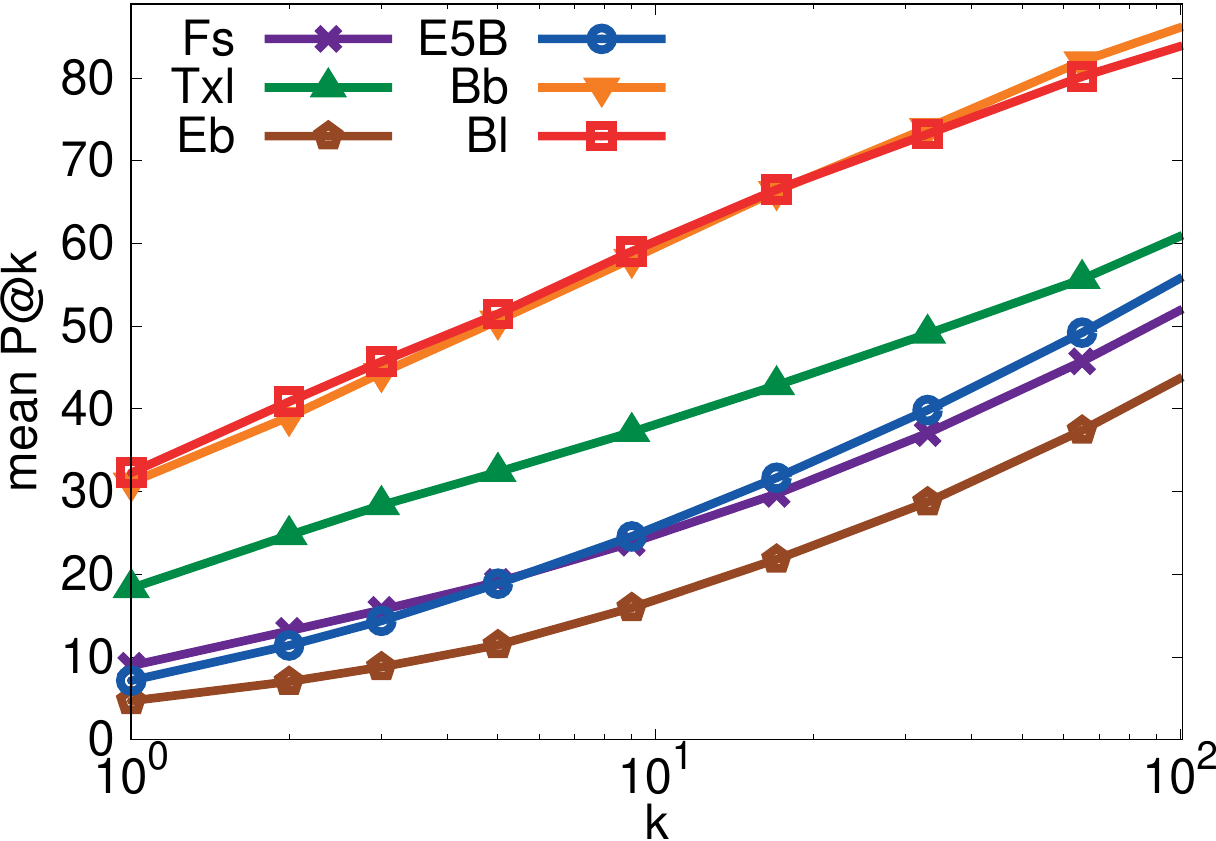}
    \caption{Mean P@k curve for T-REx varying k. Base-10 log scale for X axis.}
    \label{fig:precision}
\end{figure}

Note that a downstream model could learn to make use of knowledge in the output representations of a language model even if the correct answer is not ranked first but high enough (\ie{} a hint about the correct answer can be extracted from the output representation).
Figure \ref{fig:precision} shows the mean P@k curves for the considered models. 
For BERT, the correct object is ranked among the top ten in around 60\% of the cases and among the top 100 in 80\% of the cases.

\begin{figure}[!t]
    \centering
    \includegraphics[width=\linewidth]{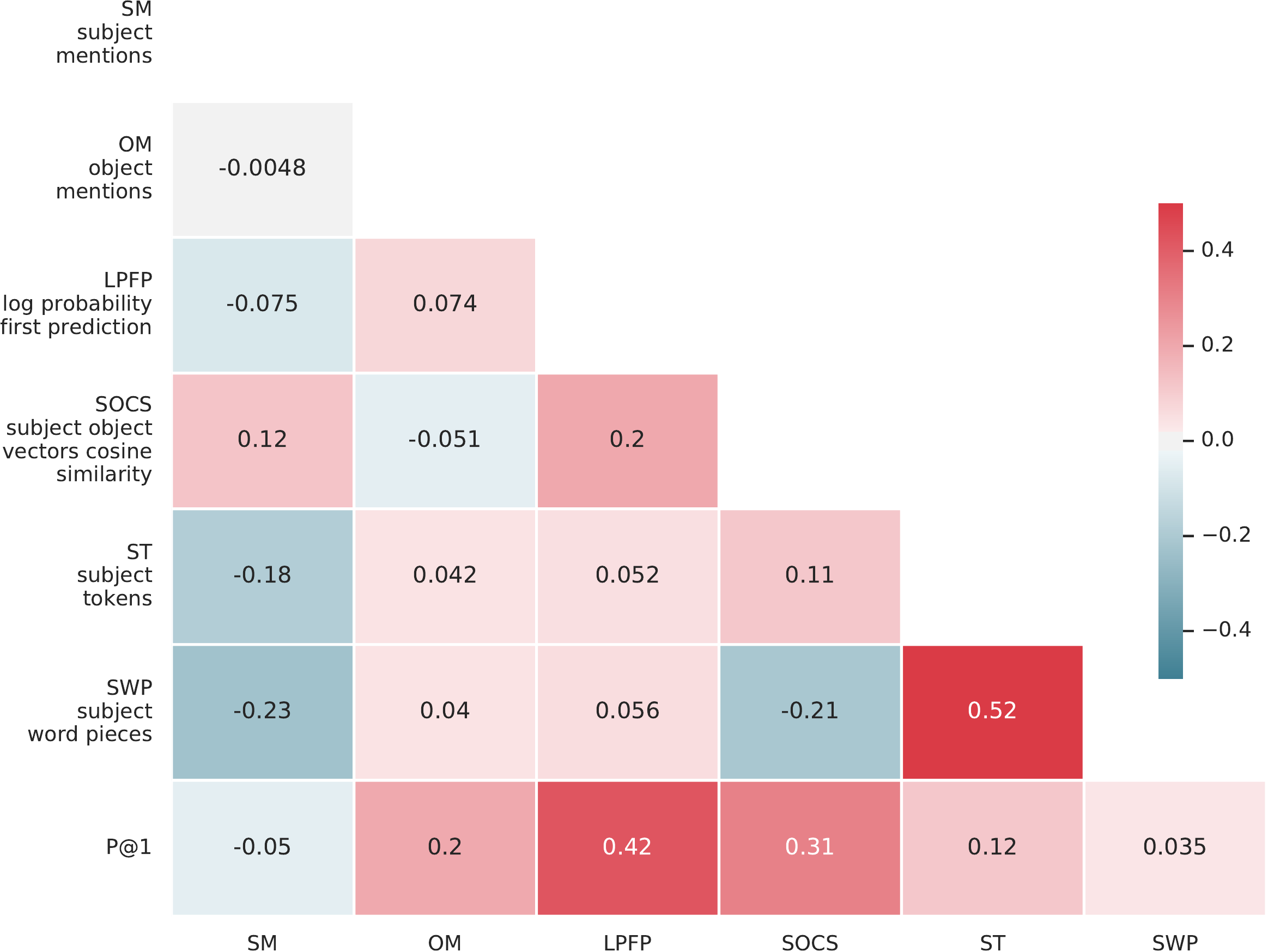}
    \caption{Pearson correlation coefficient for the P@1 of the BERT-large model on T-REx and a set of metrics: SM and OM refer to the number of times a subject and an object are mentioned in the BERT training corpus\protect\footnotemark  respectively; LPFP is the log probability score associated with the first prediction; SOCS is the cosine similarity between subject and object vectors (we use spaCy\protect\footnotemark); ST and SWP are the number of tokens in the subject with a standard tokenization and the BERT WordPiece tokenization respectively.
    }
    \label{fig:correlation}
\end{figure}

To further investigate why BERT achieves such strong results, we compute the Pearson correlation coefficient between the \textit{P@1} and a set of metrics that we report in Figure \ref{fig:correlation}. 
We notice, for instance, that the number of times an object is mentioned in the training data positively correlates with performance while the same is not true for the subject of a relation. 
Furthermore, the log probability of a prediction is strongly positively correlated with P@1.
Thus, when BERT has a high confidence in its prediction, it is often correct. 
Performance is also positively correlated with the cosine similarity between subject and object vectors, and slightly with the number of tokens in the subject.

Table \ref{tab:trexexamplesbert} shows randomly picked examples for the generation of BERT-large for cloze template queries. 
We find that BERT-large generally predicts objects of the correct type, even when the predicted object itself is not correct.

\footnotetext{The original training corpus is not available, we created our version using the same sources.}

\footnotetext{https://spacy.io}

\input{tables/TABLEEXAMPLES.tex}

To understand how the performance of a pretrained language model varies with different ways of querying for a particular fact, we analyze a maximum of 100 random facts per relation for which we randomly select 10 aligned sentences in Wikipedia from T-REx.\footnote{We exclude all facts with less than 10 alignments.} In each of the sentences, we mask the object of the fact, and ask the model to predict it. 
For several of our language models this also tests their ability to memorize and recall sentences from the training data since as the models have been trained on Wikipedia (see \Cref{tab:models}). 

\begin{figure}[!t]
    \centering
    \includegraphics[width=\linewidth]{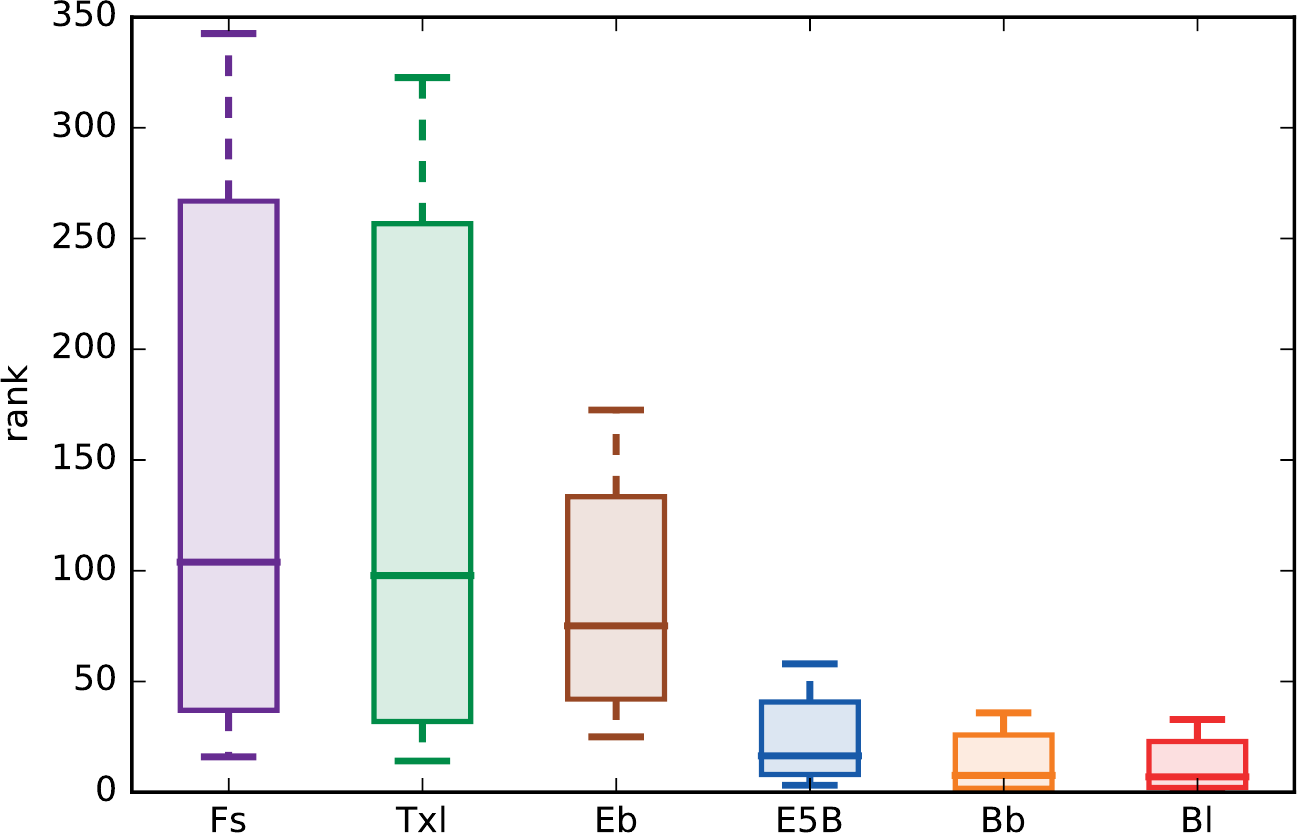}
    \caption{Average rank distribution for 10 different mentions of 100 random facts per relation in T-REx. ELMo 5.5B and both variants of BERT are least sensitive to the framing of the query but also are the most likely to have seen the query sentence during training.}
    \label{fig:mentionsbox}
\end{figure}

Figure \ref{fig:mentionsbox} shows the average distribution of the rank for ten queries per fact. 
The two BERT models and ELMo 5.5B exhibit the lowest variability while ranking the correct object close to the top on average. 
Surprisingly, the performance of ELMo original is not far from BERT, even though this model did not see Wikipedia during training. Fairseq-fconv and Transformer-XL experience a higher variability in their predictions. Note that BERT and ELMo 5.5B have been trained on a larger portion of Wikipedia than fairseq-fconv and Transformer-XL and may have seen more sentences containing the test queries during training.

\paragraph{ConceptNet}

The results on the ConceptNet corpus are in line with those reported for retrieving factual knowledge in Google-RE and T-REx.
The BERT-large model consistently achieves the best performance, and it is able to retrieve commonsense knowledge at a similar level to factual knowledge.
The lower half of Table \ref{tab:trexexamplesbert} shows generations by BERT-large for randomly sampled examples.
Some of the concepts generated by the language models are surprisingly reasonable in addition to being syntactically correct.

\paragraph{SQuAD}

Next we evaluate our system on open-domain cloze-style question answering and compare against the supervised DrQA model. Table \ref{tab:templatep1} shows a performance gap between BERT-large and the DrQA open-domain QA system on our cloze SQuAD task. 
Again, note that the pretrained language model is completely unsupervised, it is not fine-tuned, and it has no access to a dedicated information retrieval system. 
Moreover, when comparing DrQA and BERT-large in terms of P@10, we find that gap is remarkably small (57.1 for BERT-large and 63.5 for DrQA).

%% file: tables/TABLE2.tex
\begin{table*}[t!]
            \centering
            \resizebox{\textwidth}{!}{   
            \begin{tabular}{llcccccccccccc}
                \toprule
                \multirow{2}{*}{Corpus} & \multirow{2}{*}{Relation} & \multicolumn{2}{c}{Statistics} & \multicolumn{2}{|c}{Baselines} & \multicolumn{2}{c|}{KB} & \multicolumn{6}{c}{LM}  \\ 
                     & & \#Facts & \#Rel & \multicolumn{1}{|c}{Freq} & DrQA & \multicolumn{1}{c}{RE$_{n}$} &  \multicolumn{1}{c|}{RE$_{o}$} & Fs & Txl & Eb & E5B & Bb & Bl  \\ 
         \midrule
\multirow{4}{*}{Google-RE}  & \rel{birth-place} & 2937 & 1 & 4.6 & - & 3.5 & 13.8 & 4.4 & 2.7 & 5.5 & 7.5 & 14.9 & \textbf{16.1} \\ 
&\rel{birth-date} & 1825 & 1 & 1.9 & - & 0.0 & \textbf{1.9} & 0.3 & 1.1 & 0.1 & 0.1 & 1.5 & 1.4 \\ 
&\rel{death-place} & 765 & 1 & 6.8 & - & 0.1 & 7.2 & 3.0 & 0.9 & 0.3 & 1.3 & 13.1 & \textbf{14.0} \\ 
\cmidrule{2-14} 
& Total & 5527 & 3 & 4.4 & - & 1.2 & 7.6 & 2.6 & 1.6 & 2.0 & 3.0 & 9.8 & \textbf{10.5} \\
\midrule 
\multirow{4}{*}{T-REx}  & $1$-$1$ & 937 & 2 & 1.78 & - & 0.6 & 10.0 & 17.0 & 36.5 & 10.1 & 13.1 & 68.0 & \textbf{74.5} \\
&$N$-$1$ & 20006 & 23 & 23.85 & - & 5.4 & \textbf{33.8} & 6.1 & 18.0 & 3.6 & 6.5 & 32.4 & 34.2 \\
& $N$-$M$ & 13096 & 16 & 21.95 & - & 7.7 & \textbf{36.7} & 12.0 & 16.5 & 5.7 & 7.4 &  24.7 & 24.3 \\
\cmidrule{2-14} 
& Total & 34039 & 41 & 22.03 & - & 6.1 & \textbf{33.8} & 8.9 & 18.3 & 4.7 & 7.1 & 31.1 & 32.3 \\
\midrule 
ConceptNet & Total & 11458 & 16 & 4.8 & - & - & - & 3.6 & 5.7 & 6.1 & 6.2 & 15.6 & \textbf{19.2} \\
\midrule 
SQuAD & Total & 305 & - & - & \textbf{37.5} & - & - & 3.6 & 3.9 & 1.6 & 4.3 & 14.1 & 17.4 \\
\bottomrule 
\end{tabular}
}
\caption{Mean precision at one (P@1) for a frequency baseline (Freq), DrQA, a relation extraction with na\"ive entity linking (RE$_n$), oracle entity linking (RE$_o$), fairseq-fconv (Fs), Transformer-XL large (Txl),  ELMo original (Eb), ELMo 5.5B (E5B), BERT-base (Bb) and BERT-large (Bl) across the set of evaluation corpora.}
\label{tab:templatep1}
\end{table*}

%% file: tables/TABLEEXAMPLES.tex
\begin{table*}[h!]
\centering
\resizebox{\textwidth}{!}{   
\begin{tabular}{lllll}
\toprule
 & Relation & Query & Answer & Generation \\
\midrule

\multirow{23}{*}{\rotatebox[origin=c]{90}{T-Rex}} & P19 & Francesco Bartolomeo Conti was born in \underline{\hspace{2em}}. & Florence & Rome { \small [-1.8] }, \textbf{Florence { \small [-1.8] }}, Naples { \small [-1.9] }, Milan { \small [-2.4] }, Bologna { \small [-2.5] } \\

& P20 & Adolphe Adam died in \underline{\hspace{2em}}. & Paris & \textbf{Paris { \small [-0.5] }}, London { \small [-3.5] }, Vienna { \small [-3.6] }, Berlin { \small [-3.8] }, Brussels { \small [-4.0] } \\

& P279 & English bulldog is a subclass of \underline{\hspace{2em}}. & dog & dogs { \small [-0.3] }, breeds { \small [-2.2] }, \textbf{dog { \small [-2.4] }}, cattle { \small [-4.3] }, sheep { \small [-4.5] } \\

& P37 & The official language of Mauritius is \underline{\hspace{2em}}. & English & \textbf{English { \small [-0.6] }}, French { \small [-0.9] }, Arabic { \small [-6.2] }, Tamil { \small [-6.7] }, Malayalam { \small [-7.0] } \\

& P413 & Patrick Oboya plays in \underline{\hspace{2em}} position. & midfielder & centre { \small [-2.0] }, center { \small [-2.2] }, \textbf{midfielder { \small [-2.4] }}, forward { \small [-2.4] }, midfield { \small [-2.7] } \\

 & P138 & Hamburg Airport is named after \underline{\hspace{2em}}. & Hamburg & Hess { \small [-7.0] }, Hermann { \small [-7.1] }, Schmidt { \small [-7.1] }, \textbf{Hamburg { \small [-7.5] }}, Ludwig { \small [-7.5] } \\

& P364 & The original language of Mon oncle Benjamin is \underline{\hspace{2em}}. & French & \textbf{French { \small [-0.2] }}, Breton { \small [-3.3] }, English { \small [-3.8] }, Dutch { \small [-4.2] }, German { \small [-4.9] } \\

& P54 & Dani Alves plays with \underline{\hspace{2em}} . & Barcelona & Santos { \small [-2.4] }, Porto { \small [-2.5] }, Sporting { \small [-3.1] }, Brazil { \small [-3.3] }, Portugal { \small [-3.7] } \\

& P106 & Paul Toungui is a \underline{\hspace{2em}} by profession . & politician & lawyer { \small [-1.1] }, journalist { \small [-2.4] }, teacher { \small [-2.7] }, doctor { \small [-3.0] }, physician { \small [-3.7] } \\

& P527 & Sodium sulfide consists of \underline{\hspace{2em}}. & sodium & water { \small [-1.2] }, sulfur { \small [-1.7] }, \textbf{sodium { \small [-2.5] }}, zinc { \small [-2.8] }, salt { \small [-2.9] } \\

& P102 & Gordon Scholes is a member of the \underline{\hspace{2em}} political party. & Labor & Labour { \small [-1.3] }, Conservative { \small [-1.6] }, Green { \small [-2.4] }, Liberal { \small [-2.9] }, \textbf{Labor { \small [-2.9] }} \\

& P530 & Kenya maintains diplomatic relations with \underline{\hspace{2em}}. & Uganda & India { \small [-3.0] }, \textbf{Uganda { \small [-3.2] }}, Tanzania { \small [-3.5] }, China { \small [-3.6] }, Pakistan { \small [-3.6] } \\

& P176 & iPod Touch is produced by \underline{\hspace{2em}}. & Apple & \textbf{Apple { \small [-1.6] }}, Nokia { \small [-1.7] }, Sony { \small [-2.0] }, Samsung { \small [-2.6] }, Intel { \small [-3.1] } \\


& P30 & Bailey Peninsula is located in \underline{\hspace{2em}}. & Antarctica & \textbf{Antarctica { \small [-1.4] }}, Bermuda { \small [-2.2] }, Newfoundland { \small [-2.5] }, Alaska { \small [-2.7] }, Canada { \small [-3.1] } \\

& P178 & JDK is developed by \underline{\hspace{2em}}. & Oracle & IBM { \small [-2.0] }, Intel { \small [-2.3] }, Microsoft { \small [-2.5] }, HP { \small [-3.4] }, Nokia { \small [-3.5] } \\


 & P1412 & Carl III used to communicate in \underline{\hspace{2em}}. & Swedish & German { \small [-1.6] }, Latin { \small [-1.9] }, French { \small [-2.4] }, English { \small [-3.0] }, Spanish { \small [-3.0] } \\

& P17 & Sunshine Coast, British Columbia is located in \underline{\hspace{2em}}. & Canada & \textbf{Canada { \small [-1.2] }}, Alberta { \small [-2.8] }, Yukon { \small [-2.9] }, Labrador { \small [-3.4] }, Victoria { \small [-3.4] } \\

& P39 & Pope Clement VII has the position of \underline{\hspace{2em}} . & pope & cardinal { \small [-2.4] }, Pope { \small [-2.5] }, \textbf{pope { \small [-2.6] }}, President { \small [-3.1] }, Chancellor { \small [-3.2] } \\

& P264 & Joe Cocker is represented by music label \underline{\hspace{2em}}. & Capitol & EMI { \small [-2.6] }, BMG { \small [-2.6] }, Universal { \small [-2.8] }, \textbf{Capitol { \small [-3.2] }}, Columbia { \small [-3.3] } \\

& P276 & London Jazz Festival is located in \underline{\hspace{2em}}. & London & \textbf{London { \small [-0.3] }}, Greenwich { \small [-3.2] }, Chelsea { \small [-4.0] }, Camden { \small [-4.6] }, Stratford { \small [-4.8] } \\

& P127 & Border TV is owned by \underline{\hspace{2em}}. & ITV & Sky { \small [-3.1] }, \textbf{ITV { \small [-3.3] }}, Global { \small [-3.4] }, Frontier { \small [-4.1] }, Disney { \small [-4.3] } \\

& P103 & The native language of Mammootty is \underline{\hspace{2em}}. & Malayalam & \textbf{Malayalam { \small [-0.2] }}, Tamil { \small [-2.1] }, Telugu { \small [-4.8] }, English { \small [-5.2] }, Hindi { \small [-5.6] } \\

& P495 & The Sharon Cuneta Show was created in \underline{\hspace{2em}}. & Philippines & Manila { \small [-3.2] }, \textbf{Philippines { \small [-3.6] }}, February { \small [-3.7] }, December { \small [-3.8] }, Argentina { \small [-4.0] } \\

\midrule

\multirow{10}{*}{\rotatebox[origin=c]{90}{ConceptNet}} & AtLocation & You are likely to find a overflow in a \underline{\hspace{2em}}. & drain & sewer { \small [-3.1] }, canal { \small [-3.2] }, toilet { \small [-3.3] }, stream { \small [-3.6] }, \textbf{drain { \small [-3.6] }} \\

& CapableOf & Ravens can \underline{\hspace{2em}}. & fly & \textbf{fly { \small [-1.5] }}, fight { \small [-1.8] }, kill { \small [-2.2] }, die { \small [-3.2] }, hunt { \small [-3.4] } \\

& CausesDesire & Joke would make you want to \underline{\hspace{2em}}. & laugh & cry { \small [-1.7] }, die { \small [-1.7] }, \textbf{laugh { \small [-2.0] }}, vomit { \small [-2.6] }, scream { \small [-2.6] } \\

& Causes & Sometimes virus causes \underline{\hspace{2em}}. & infection & disease { \small [-1.2] }, cancer { \small [-2.0] }, \textbf{infection { \small [-2.6] }}, plague { \small [-3.3] }, fever { \small [-3.4] } \\

& HasA & Birds have \underline{\hspace{2em}}. & feathers & wings { \small [-1.8] }, nests { \small [-3.1] }, \textbf{feathers { \small [-3.2] }}, died { \small [-3.7] }, eggs { \small [-3.9] } \\

& HasPrerequisite & Typing requires \underline{\hspace{2em}}. & speed & patience { \small [-3.5] }, precision { \small [-3.6] }, registration { \small [-3.8] }, accuracy { \small [-4.0] }, \textbf{speed { \small [-4.1] }} \\

& HasProperty & Time is \underline{\hspace{2em}}. & finite & short { \small [-1.7] }, passing { \small [-1.8] }, precious { \small [-2.9] }, irrelevant { \small [-3.2] }, gone { \small [-4.0] } \\

& MotivatedByGoal & You would celebrate because you are \underline{\hspace{2em}}. & alive & happy { \small [-2.4] }, human { \small [-3.3] }, \textbf{alive { \small [-3.3] }}, young { \small [-3.6] }, free { \small [-3.9] } \\

& ReceivesAction & Skills can be \underline{\hspace{2em}}. & taught & acquired { \small [-2.5] }, useful { \small [-2.5] }, learned { \small [-2.8] }, combined { \small [-3.9] }, varied { \small [-3.9] } \\

& UsedFor & A pond is for \underline{\hspace{2em}}. & fish & swimming { \small [-1.3] }, fishing { \small [-1.4] }, bathing { \small [-2.0] }, \textbf{fish { \small [-2.8] }}, recreation { \small [-3.1] } \\

\bottomrule 
\end{tabular}
}
\caption{Examples of generation for BERT-large. The last column reports the top five tokens generated together with the associated log probability (in square brackets).}
\label{tab:trexexamplesbert}
\end{table*}

%% file: 6_conclusion.tex
\section{Discussion and Conclusion}
We presented a systematic analysis of the factual and commonsense knowledge in publicly available pretrained language models \emph{as is} and found that BERT-large is able to recall such knowledge better than its competitors and at a level remarkably competitive with non-neural and supervised alternatives.
Note that we did \emph{not} compare the ability of the corresponding architectures and objectives to capture knowledge in a given body of text but rather focused on the knowledge present in the weights of existing pretrained models that are being used as starting points for many researchers' work. Understanding which aspects of data our commonly-used models and learning algorithms are capturing is a crucial field of research and this paper complements the many studies focused on the learned linguistic properties of the data.

We found that it is non-trivial to extract a knowledge base from text that performs on par to directly using pretrained BERT-large. 
This is despite providing our relation extraction baseline with only data that is likely expressing target facts, thus reducing potential for false negatives, as well as using a generous entity-linking oracle.
We suspected BERT might have an advantage due to the larger amount of data it has processed, so we added Wikitext-103 as additional data to the relation extraction system and observed no significant change in performance.
This suggests that while relation extraction performance might be difficult to improve with more data, language models trained on ever growing corpora might become a viable alternative to traditional knowledge bases extracted from text in the future.

In addition to testing future pretrained language models using the LAMA probe, we are interested in quantifying the variance of recalling factual knowledge with respect to varying natural language templates. Moreover, assessing multi-token answers remains an open challenge for our evaluation setup.